%% file: main.tex
\documentclass[10pt,twocolumn]{article}

\usepackage{times}
\usepackage{graphicx}
\usepackage{amsmath,amssymb}
\usepackage{cite}         
\usepackage[pagebackref,breaklinks,colorlinks]{hyperref}

\input{preamble}

\title{Beyond Visual Similarity: Rule-Guided Multimodal Clustering with explicit domain rules}

\author{Kishor Datta Gupta\\
Department of Cyber Physical Systems\\
Clark Atlanta University, Atlanta, GA, USA\\
{\tt\small kgupta@cau.edu}
\and
Mohd Ariful Haque\\
Department of Cyber Physical Systems\\
Clark Atlanta University, Atlanta, GA, USA\\
{\tt\small mohdariful.haque@students.cau.edu}
\and
Marufa Kamal\\
Department of Computer Science and Engineering\\
BRAC University, Dhaka, Bangladesh\\
{\tt\small marufa.kamal1@g.bracu.ac.bd}
\and
Ahmed Rafi Hasan\\
Department of Computer Science and Engineering\\
United International University, Dhaka, Bangladesh\\
{\tt\small ahasan191131@bscse.uiu.ac.bd}
\and
Md. Mahfuzur Rahman\\
Department of Cyber Physical Systems\\
Clark Atlanta University, Atlanta, GA, USA\\
{\tt\small mdmahfuzur.rahman@students.cau.edu}
\and
Roy George\\
Department of Cyber Physical Systems\\
Clark Atlanta University, Atlanta, GA, USA\\
{\tt\small rgeorge@cau.edu}
}

\begin{document}
\maketitle
\input{sec/0_abstract}    
\input{sec/1_intro}
\input{sec/2_literature_review}
\input{sec/3_methodology}
\input{sec/4_dataset}
\input{sec/5_results_analysis}
\input{sec/6_conclusion}

{
    \small
    \bibliographystyle{IEEEtran.bst}
    \bibliography{main}
}

\end{document}

%% file: preamble.tex
\usepackage{subcaption}
\usepackage{booktabs}

\usepackage{listings}
\usepackage{mathtools}

\usepackage{algorithm}
\usepackage{algpseudocode}

\usepackage{multirow}

%% file: sec/0_abstract.tex
\begin{abstract}
Traditional clustering techniques often rely solely on similarity in the input data, limiting their ability to capture structural or semantic constraints that are critical in many domains. We introduce the Domain-Aware Rule-Triggered Variational Autoencoder (DART-VAE), a rule-guided multimodal clustering framework that incorporates domain-specific constraints directly into the representation learning process. DART-VAE extends the VAE architecture by embedding explicit rules, semantic representations, and data-driven features into a unified latent space, while enforcing constraint compliance through rule-consistency and violation penalties in the loss function. Unlike conventional clustering methods that rely only on visual similarity or apply rules as post-hoc filters, DART-VAE treats rules as first-class learning signals. The rules are generated by LLMs, structured into knowledge graphs, and enforced through a loss function combining reconstruction, KL divergence, consistency, and violation penalties. Experiments on aircraft and automotive datasets demonstrate that rule-guided clustering produces more operationally meaningful and interpretable clusters—for example, isolating UAVs, unifying stealth aircraft, or separating SUVs from sedans—while improving traditional clustering metrics. However, the framework faces challenges: LLM-generated rules may hallucinate or conflict, excessive rules risk overfitting, and scaling to complex domains increases computational and consistency difficulties. By combining rule encodings with learned representations, DART-VAE achieves more meaningful and consistent clustering outcomes than purely data-driven models, highlighting the utility of constraint-guided multimodal clustering for complex, knowledge-intensive settings.
\end{abstract}

%% file: sec/1_intro.tex
\section{Introduction}
\label{sec:intro}

Many visual clustering methodologies presume that visual similarity reflects functional similarity; however, appearance and function can diverge in specialized domains. General image clustering methods perform well in natural image domains~\cite{10.1007/978-3-030-58607-2_16}, yet frequently falter in specialized datasets where operational semantics precede visual appearance. Public benchmark models, while achieving impressive results on large-scale datasets, often fail to transfer effectively when fine-tuned for domain-specific tasks. Vision Transformers (ViTs) have shown promise in this space due to their strong representational capacity, but their success typically relies on very large volumes of training data. Attempting to fine-tune ViTs with limited specialized datasets often results in overfitting\cite{liu2021efficient, zhao2025missing}, whereas relying solely on public datasets without proper adaptation leads to underfitting and poor generalization in operational settings. Multimodal techniques~\cite{radford2021learning, jia2021scaling} attempt to bridge this gap by inferring relationships through large-scale data-driven optimization; however, without embedding domain-specific constraints, they risk overlooking the expert knowledge essential for high-stakes applications. 

As an illustrative case, stealth bombers and fighters often adopt angular geometries to minimize radar cross-section, yet aircraft with equivalent functions may present markedly distinct visual profiles due to differing manufacturer philosophies and generational design shifts. Similarly, crossover SUVs may appear visually similar, but subtle distinctions in structure and purpose reflect their alignment with separate market segments. These examples highlight that visual similarity alone is insufficient to capture functional or operational equivalence.

\begin{figure*}[!ht]
    \centering
    \includegraphics[width=0.7\linewidth]{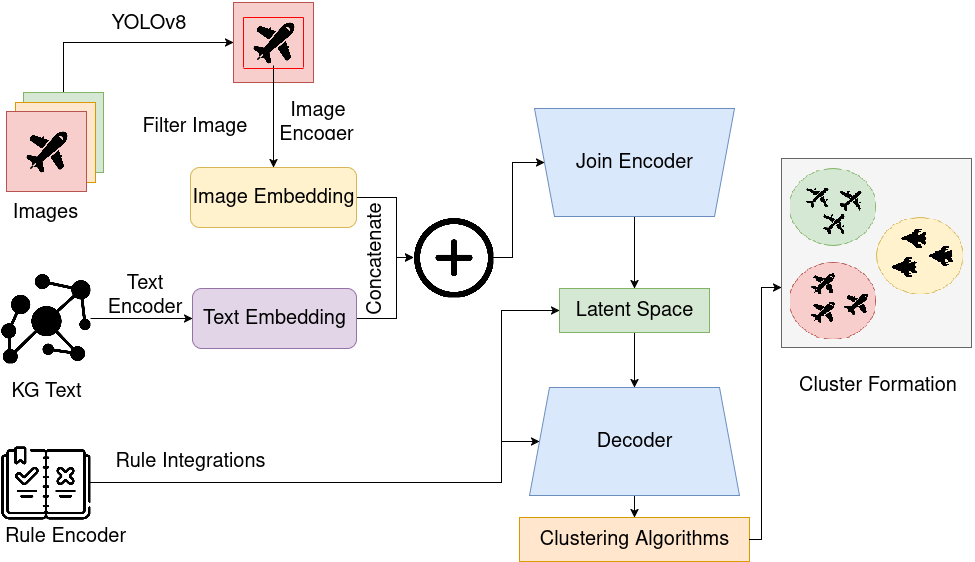}
    \caption{DART-VAE: Domain-Aware Rule-Guided Training for Variational Autoencoders. Our framework integrates visual features, semantic knowledge from structured ontologies, and explicit domain rules (Generated by LLM) through specialized encoders, ensuring learned representations inherently respect operational constraints.}
    \label{fig:dart_vae}
\end{figure*}

Our primary contributions:
    \begin{itemize} 
    
        \item A multimodal VAE architecture encodes visual features, semantic knowledge graphs, and explicit domain rules(Generated by LLM) via specialized pathways, ensuring that learned representations comply with operational constraints.
        
        \item  A multifaceted objective function that equilibrates reconstruction fidelity with rule enforcement through the integration of consistency and violation losses, thereby maintaining generative properties while conforming to expert-defined relationships.
        
    \end {itemize}

 This study formulates a systematic framework for integrating expert knowledge into deep clustering, relevant to domains such as medical imaging, industrial inspection, and scientific analysis, where domain constraints are paramount.

%% file: sec/2_literature_review.tex
\section{Related Works}
\label{sec:related}
Clustering techniques for visual data have achieved notable success in grouping images based on surface-level similarities. Methods such as k-means, fuzzy c-means, Deep Embedded Clustering (DEC)~\cite{xie2016unsupervised}, and Variational Deep Embedding (VaDE)~\cite{jiang2016variational} optimize purely data-driven objectives but often fail in capturing fine-grained, domain-specific distinctions. In specialized datasets (e.g., vehicles), these approaches neglect semantic and structural knowledge required for meaningful subclass separation. While joint feature learning frameworks such as DAC~\cite{chang2017deep} and its refinements with local structure preservation~\cite{guo2017improved} and augmentation~\cite{guo2018deep} improve robustness, they remain limited to visual similarity. More recent pipelines, including CPP~\cite{chu2024image}, leverage pre-trained features for scalability yet still lack explicit integration of domain constraints. Constraint-based clustering has introduced must-link and cannot-link supervision~\cite{ge2007constraint}, and generative models have been applied to encode expert priors~\cite{andreeva2020catalysis}. Multimodal approaches enhance interpretability by aligning visual and semantic features~\cite{chen2021multimodal}, and fine-grained methods have advanced through diffusion-based~\cite{yang2024dific} and bipartite factorization~\cite{peng2024fine}. Despite these efforts, existing methods rarely embed domain rules as first-class constraints during representation learning, leaving a gap for frameworks that unify visual, semantic, and rule-based guidance.

\begin{table*}[!htbp]
\centering
\caption{Rule-Guided Clustering Improvement Examples}
\label{tab:improvements}
\renewcommand{\arraystretch}{0.9}
\setlength{\tabcolsep}{6pt}
\begin{tabular}{p{4cm} p{4cm} p{3cm} p{3.8cm}}
\toprule
\textbf{Example} & \textbf{Without Rules} & \textbf{With Rules} & \textbf{Improvement} \\
\midrule
\multicolumn{4}{l}{\textbf{\textit{Aircraft Domain}}} \\
MQ-9 Reaper (UAV) & Mixed with F-18 fighter & Isolated UAV cluster & $\checkmark$ UAV separation \\
F-22, Su-57 (Stealth) & Scattered over multiple clusters & Unified stealth cluster & $\checkmark$ Technology consistency \\
C-130, C-2 (Transport) & Mixed with A-10 combat  & Pure transport cluster & $\checkmark$ Mission separation \\
\midrule
\multicolumn{4}{l}{\textbf{\textit{Vehicle Domain}}} \\
BMW M3 (Performance) & Mixed with Toyota Camry & Performance cluster & $\checkmark$ Performance separation \\
Ferrari 488 (Luxury) & Grouped with Honda Civic & Luxury sports cluster & $\checkmark$ Market segmentation \\
Range Rover (SUV) & Mixed with sedan vehicles & SUV cluster & $\checkmark$ Body style coherence \\
\bottomrule
\end{tabular}
\end{table*}

%% file: sec/3_methodology.tex
\section{Methodology}
\label{sec:methodology}

\subsection{Problem Formulation and Motivation}

The primary challenge in specialized domain clustering is the disparity between visual appearance and functional purpose. Conventional clustering techniques depend solely on pixel-level or derived visual features, operating under the premise that visually analogous objects possess functional similarities. This assumption fails profoundly in areas where form adheres to highly specialized function rather than aesthetic resemblance.

\textbf{Human vs. Machine Perception Disparity}: Humans possess inherent domain knowledge enabling them to instantly differentiate between a combat fighter and a transport aircraft, despite their similar visual characteristics. Machine learning models, however, lack contextual comprehension and rely solely on superficial visual patterns. For example, the F-16 Fighting Falcon and C-130 Hercules may seem analogous in aerial images due to their monoplane design; however, they fulfill distinctly different operational roles—one as an air superiority fighter and the other as a tactical transport aircraft.

\textbf{Operational Reality}: In military aviation, aircraft with nearly identical visual signatures can serve very different functions. The F-22 Raptor and F-117 Nighthawk both possess angular, stealth-optimized designs for radar evasion; however, the F-22 serves as an air superiority fighter with supercruise capability, whereas the F-117 operated as a precision strike bomber. Conversely, functionally analogous aircraft may exhibit significant visual diversity owing to varying design epochs, manufacturers, and technological methodologies.

\textbf{Clustering Inadequacy}: Conventional clustering algorithms consistently categorize the MQ-9 Reaper (UAV), F-18 Hornet (fighter), and KC-135 Stratotanker (refueling aircraft) solely based on visual resemblance, resulting in operationally irrelevant clusters that contravene essential military doctrine principles.

Our DART-VAE framework addresses this limitation by embedding domain-specific physical rules (which generated by LLM and Domain specific Books) directly into the representation learning process, ensuring that learned embeddings respect both visual coherence and operational semantics. We define this as the acquisition of a latent representation $z \in \mathbb{R}^d$ that organizes data points based on visual similarity and domain constraints $R = \{r_j\}_{j=1}^M$.

\subsection{Overall Architecture}

The DART-VAE framework employs a three-stage pipeline that methodically converts raw multimodal data into constraint-aware latent representations appropriate for domain-informed clustering.

\textbf{Stage 1: Multimodal Feature Extraction} Raw images are subjected to object detection via YOLOv8 to delineate regions of interest (ROI) with adaptive padding, thereby removing background noise that results in erroneous groupings. Concurrently, structured domain knowledge from JSON-formatted knowledge graphs (aircraft) or CSV metadata (vehicles) is processed using Sentence-BERT to produce semantic embeddings. These Knowledge graphs and metadata are acquire by LLM fine-tuned by domain-specific contents). Binary rule features are extracted and encoded via specialized MLPs.

\textbf{Stage 2: Constraint-Guided Representation Learning} The DART-VAE encoder processes concatenated multimodal features (visual + semantic + rules) through a joint encoder network that learns the posterior distribution $q(z|x,t,r)$. The formation of latent space is directed by a multi-faceted loss function that equilibrates reconstruction accuracy, KL regularization, rule adherence, and penalty for violations.

\textbf{Stage 3: Rule-Validated Clustering} Acquired latent representations undergo hard (K-means) and soft (Fuzzy C-means) clustering, followed by rule-guided refinement, in which constraint violations lead to reassignment to the nearest compliant cluster, based on both latent distance and rule adherence.


\subsection{Domain-Specific Physical Rules}

The core innovation of DART-VAE lies in the explicit formalization of domain knowledge as enforceable constraints. We establish unique rule sets for each domain that encapsulate essential operational principles.

\begin{algorithm}[!ht]
\caption{DART-VAE: Domain-Aware Rule-Triggered Clustering}
\label{alg:dart_vae}
\begin{algorithmic}[1]
\Require Multimodal dataset $\mathcal{D} = \{(x_i, t_i, r_i)\}_{i=1}^N$, Domain rules $\mathcal{R}$
\Ensure Rule-compliant clusters $\mathcal{C} = \{C_k\}_{k=1}^K$

\State \textbf{Stage 1: Multimodal Feature Extraction}
\State $X_{\text{roi}} \leftarrow \text{ObjectDetection}(\{x_i\})$ \Comment{ROI extraction}
\State $F_v \leftarrow \text{VisualEncoder}(X_{\text{roi}})$ \Comment{Visual features}
\State $F_t \leftarrow \text{SemanticEncoder}(\{t_i\})$ \Comment{Knowledge features}
\State $F_r \leftarrow \text{RuleEncoder}(\{r_i\}, \mathcal{R})$ \Comment{Rule features}
\State $F_{\text{joint}} \leftarrow \text{Concatenate}(F_v, F_t, F_r)$

\State \textbf{Stage 2: Constraint-Guided Representation Learning}
\For{epoch $e = 1$ to $T$}
    \State $\mu, \sigma^2 \leftarrow \text{Encoder}(F_{\text{joint}})$
    \State $Z \leftarrow \text{Reparameterize}(\mu, \sigma^2)$ \Comment{Latent sampling}
    \State $\mathcal{L} \leftarrow \mathcal{L}_{\text{recon}} + \beta\mathcal{L}_{\text{KL}} + \alpha_e(\mathcal{L}_{\text{consistency}} + \mathcal{L}_{\text{violation}})$
    \State Update $\theta$ via $\nabla_\theta \mathcal{L}$ \Comment{Progressive rule integration}
\EndFor

\State \textbf{Stage 3: Rule-Validated Clustering}
\State $\mathcal{C}_{\text{init}} \leftarrow \text{Clustering}(Z)$ \Comment{K-means or Fuzzy C-means}
\For{each constraint $r \in \mathcal{R}$}
    \For{each cluster $C_k \in \mathcal{C}_{\text{init}}$}
        \If{$\text{ViolatesRule}(C_k, r)$}
            \State Reassign violating samples to nearest compliant clusters
        \EndIf
    \EndFor
\EndFor
\State \Return $\mathcal{C}$
\end{algorithmic}
\end{algorithm}

\subsubsection{Aircraft Domain Rules}

According to military aviation doctrine and aerospace engineering principles, a fine-tuned LLM generated four essential constraints:

\textbf{Rule 1: Stealth Technology Consistency} Stealth aircraft represent a highly specialized technological category requiring sophisticated systems integration. Aircraft with stealth capabilities must exhibit technological reliability via advanced avionics systems and possess either air superiority (fighter classification) or supersonic cruise capability.

\textit{Formal Definition}: $\forall a \in \text{Aircraft}: \text{is\_stealth}(a) \rightarrow (\text{has\_advanced\_avionics}(a) \land (\text{is\_fighter}(a) \lor \text{has\_supercruise}(a)))$

\textit{Physical Rationale}: Stealth technology necessitates advanced radar systems, electronic warfare capabilities, and intricate flight controls. The F-22 Raptor integrates stealth capabilities with supercruise, whereas the F-117 Nighthawk depended on sophisticated avionics for precision strike operations.

\textbf{Rule 2: UAV Operational Separation} Unmanned and manned aircraft function under distinct doctrines, certification criteria, and operational protocols. They must uphold distinct clustering boundaries to accurately represent operational reality.

\textit{Formal Definition}: $\forall a_i, a_j \in C_k: \text{is\_uav}(a_i) \leftrightarrow \text{is\_uav}(a_j)$

\textit{Physical Rationale}: UAVs such as the MQ-9 Reaper function with different risk profiles, endurance capacities, and mission specifications in contrast to manned aircraft like the F-16.

\textbf{Rule 3: Mission-Type Doctrinal Enforcement} Military aircraft are designed and optimized for specific mission profiles. Combat platforms must not be grouped with transport or logistics aircraft due to inherent disparities in operational requirements, threat environments, and deployment patterns.

\textit{Formal Definition}: $\forall a_i, a_j \in C_k: \text{mission\_type}(a_i) = \text{combat} \rightarrow \text{mission\_type}(a_j) \neq \text{transport}$

\textit{Physical Rationale}: Combat aircraft like the A-10 Thunderbolt II are armored for survivability in hostile environments, while transport aircraft like the C-130 prioritize cargo capacity and operational versatility.

\textbf{Rule 4: Physical Attribute Coherence} Aircraft within the same operational cluster must demonstrate comparable fundamental physical attributes, including propulsion systems and performance envelopes, indicative of analogous operational requirements.

\textit{Formal Definition}: $\forall a_i, a_j \in C_k: \text{engine\_type}(a_i) = \text{engine\_type}(a_j) \land \text{speed\_class}(a_i) = \text{speed\_class}(a_j)$

\textit{Physical Rationale}: Turbofan-powered aircraft function within individual performance parameters compared to turboprop aircraft, influencing range, altitude capabilities, and mission appropriateness.

\subsubsection{Automotive Domain Rules}

For vehicles, OPENAI GPT3.5 outlines four rules that encapsulate market segmentation and engineering principles for vehicles.

\textbf{Rule 1: Body Style Coherence} Vehicles with fundamentally diverse body structures cater to specific market niches and usage patterns.  SUVs, sedans, and convertibles cater to their own market requirements and should maintain cluster differentiation.

\textbf{Rule 2: Performance Tier Consistency} Economy cars and high-performance cars have different engineering aims and target various market segments.  Performance cars put handling and power-to-weight ratios first, whereas economy cars maximize cost and fuel efficiency.

\textbf{Rule 3: Dimensional Proportionality} Vehicles with notably distinct physical proportions (height-to-length ratios, ground clearance) fulfill diverse practical functions and should remain distinct in clustering.

\textbf{Rule 4: Luxury Market Segmentation} Luxury and standard market vehicles represent distinct value propositions with different feature sets, pricing strategies, and brand positioning.

\subsection{Multimodal Feature Extraction Pipeline}

\textbf{Visual Feature Processing}: We utilize YOLOv8 for object detection and ROI extraction, implementing adaptive padding to encompass entire aircraft and vehicle structures while minimizing background noise.  Our experimental validation demonstrates that background contamination substantially affects clustering quality, rendering ROI extraction an essential preprocessing step.

\textit{Aircraft Domain}: Three convolutional layers Conv2d(3,32) $\rightarrow$ Conv2d(32,64) $\rightarrow$ Conv2d(64,128), flattening yields 100,352D feature vectors (128 $\times$ 28 $\times$ 28), 20\% adaptive padding applied during ROI extraction.

\textit{Automotive Domain}: Four convolutional layers with adaptive average pooling, culminating in a 256-dimensional visual feature representation, with a confidence threshold of 0.25 for vehicle detection.

\textbf{Semantic Knowledge Encoding}: Sentence-BERT (all-mpnet-base-v2) processes structured domain expertise.  Aircraft use JSON-formatted knowledge graph triples with technical specifications, operational roles, and performance attributes contained in 384D vectors and compressed to 256D via 2-layer MLP.  Sentence-BERT to 768D embeddings procedure CSV brand hierarchies, technical specifications, body styles, and market categories for vehicles.

\textbf{Rule Feature Engineering}: Domain constraints are implemented as explicit feature vectors via specialized MLPs.  Aircraft utilize 10 binary attributes (is\_stealth, is\_uav, has\_crew, has\_supercruise, has\_advanced\_avionics, mission\_type indications) processed using MLP to 16-dimensional space.  Vehicles utilize 18 derived attributes from the four automotive rules (body\_style\_category, performance\_tier, size\_ratios, luxury\_indicators) compressed into a 32-dimensional space.

\subsection{DART-VAE Architecture Details}

\textbf{Joint Encoding}: The multimodal feature fusion creates domain-specific joint representations. Aircraft: $f_{\text{joint}} \in \mathbb{R}^{100,624}$ (visual: 100,352D + semantic: 256D + rules: 16D). Vehicles: $f_{\text{joint}} \in \mathbb{R}^{100,388}$ (visual: 100,352D + semantic: 768D + rules: 32D).

\textbf{Encoder Network}: A progressive compression architecture maps joint features to latent parameters: $h_1 = \text{ReLU}(\text{Linear}(f_{\text{joint}}, 512))$, $h_2 = \text{ReLU}(\text{Linear}(h_1, 256))$, $\mu, \log \sigma^2 = \text{Linear}(h_2, 64), \text{Linear}(h_2, 64)$.

\textbf{Latent Sampling}: The reparameterization trick enables gradient-based optimization: $z = \mu + \sigma \odot \epsilon$, where $\epsilon \sim \mathcal{N}(0, I_{64})$.

\textbf{Decoder Network}: Reconstructs multimodal features to ensure representation fidelity: $h_3 = \text{ReLU}(\text{Linear}(z, 256))$, $h_4 = \text{ReLU}(\text{Linear}(h_3, 512))$, $f_{\text{reconstructed}} = \text{Linear}(h_4, \text{dim}(f_{\text{joint}}))$.

\subsection{Multi-Component Loss Function}
\label{sec:loss_function}

Our training objective combines traditional VAE losses with rule-specific penalties that address the fundamental challenge observed in our aircraft and automotive clustering experiments:

\begin{equation}
\mathcal{L}_{total} = \mathcal{L}_{recon} + \beta\mathcal{L}_{KL} + \alpha(\mathcal{L}_{consistency} + \mathcal{L}_{violation})
\label{eq:total_loss}
\end{equation}

\textbf{Reconstruction Loss} ($\mathcal{L}_{recon}$): Standard VAE reconstruction ensures that our multimodal features; visual ROI features, knowledge graph embeddings, and rule encodings, can be faithfully recovered from the latent space:
\begin{equation}
\mathcal{L}_{recon} = \text{MSE}(f_{joint}, \hat{f}_{joint})
\label{eq:recon_loss}
\end{equation}

\textbf{KL Divergence Loss} ($\mathcal{L}_{KL}$): The standard VAE regularization prevents latent space collapse:
\begin{equation}
\mathcal{L}_{KL} = -\frac{1}{2} \sum_{i=1}^{d} (1 + \log \sigma_i^2 - \mu_i^2 - \sigma_i^2)
\label{eq:kl_loss}
\end{equation}

\textbf{Rule Consistency Loss} ($\mathcal{L}_{consistency}$): This component indicates a key finding from our experiments: aircraft with analogous operational profiles should exhibit comparable latent representations, despite visual differences.
\begin{equation}
\mathcal{L}_{consistency} = \sum_{i,j} \text{MSE}(\text{sim}(z_i, z_j), \text{sim}(r_i, r_j))
\label{eq:consistency_loss}
\end{equation}

The function $\text{sim}(r_i, r_j)$ estimates cosine similarity between 16-dimensional rule features generated by the rule encoder, rather than raw binary inputs.  This groups the MQ-9 and TB2 UAVs despite their visual profiles and aligns BMW and Mercedes premium automobiles in latent space despite brand distinctions.

\textbf{Rule Violation Loss} ($\mathcal{L}_{violation}$): Direct constraint enforcement emerges from our domain analysis:
\begin{equation}
\mathcal{L}_{violation} = \text{MSE}(\sigma(v_{pred}), v_{target})
\label{eq:violation_loss}
\end{equation}

where $v_{pred} \in \mathbb{R}^{N \times 4}$ are the raw violation predictions from the rule predictor network, $\sigma$ is the sigmoid activation, and $v_{target} \in \{0,1\}^{N \times 4}$ are the binary violation targets computed from the logical rules.

\textbf{Rule Weight Configuration}: Reconstruction fidelity and domain constraint enforcement are balanced using a 0.15 rule weight in airplane (40 epochs) and automobile (30 epochs) domain training.

%% file: sec/4_dataset.tex
\section{Experiments}
\label{sec:dataset}

\subsection{Datasets and Preprocessing}

\textbf{Military Aircraft}: We conducted a two-stage data cleansing process on 8,311 raw photos from the military aircraft dataset~\cite{a2015_military_aircraft_detection}. Preliminary filtering employing Mask R-CNN, succeeded by YOLOv8~\cite{yolov8_ultralytics} with dual parameters (confidence $\geq$ 0.97 and aircraft coverage $\geq$ 60\%), produced 3,103 high-quality images encompassing 77 aircraft types. To ensure computational efficiency and uniform assessment across rule configurations, we chose a representative subset of 800 photos (after YOLOv8 filtering, preserving 728 images with aircraft detections) encompassing 76 aircraft classes for our clustering tests. Every aircraft is marked with operating characteristics and connected to a JSON knowledge graph that includes technical specifications as triples.

\noindent \textbf{Automotive Dataset}: The vehicle dataset~\cite{unit293_car_models_3887} was subjected to quality control comparable to aircraft photos, using YOLOv8 to exclude non-vehicle images, notably interiors. A hierarchical brand/model/year directory structure and CSV metadata with technical specifications, body styles, segments, and dimensional attributes from manufacturer databases comprise the cleaned dataset.

\subsection{Implementation Details}
All tests utilize the PyTorch framework using NVIDIA RTX 6000 Ada Generation GPUs, each equipped with 49GB of VRAM. The visual encoder utilizes a custom CNN architecture with domain-specific configurations, comprising three convolutional layers for aircraft and four for cars. Text encoding using Sentence-BERT (all-mpnet-base-v2). In the preprocessing phase, YOLOv8 does object detection with a confidence threshold of 0.25 and adaptive padding for region of interest extraction. Optimization employs AdamW with a weight decay of $\lambda = 10^{-4}$. Training setups are specialized to their domains: airplane experiments utilize 40 epochs across all rule configurations, whereas car experiments employ 30 epochs with a 4-rule configuration.



\begin{figure*}[!htbp]
  \centering
   \begin{subfigure}{0.33\textwidth}
    \centering
    \includegraphics[width=\textwidth]{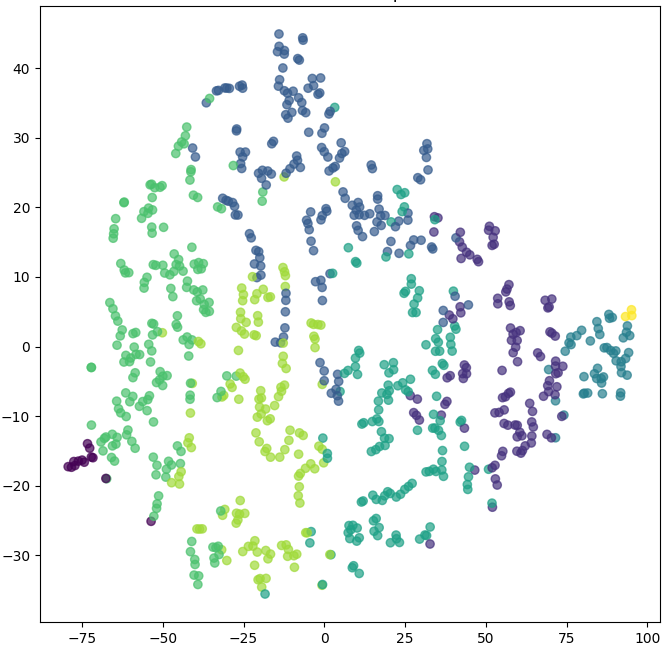}
    \caption{Hard clustering without rules}
    \label{fig:2rules_aircraft}
  \end{subfigure}
    \begin{subfigure}{0.33\textwidth}
    \centering
    \includegraphics[width=\textwidth]{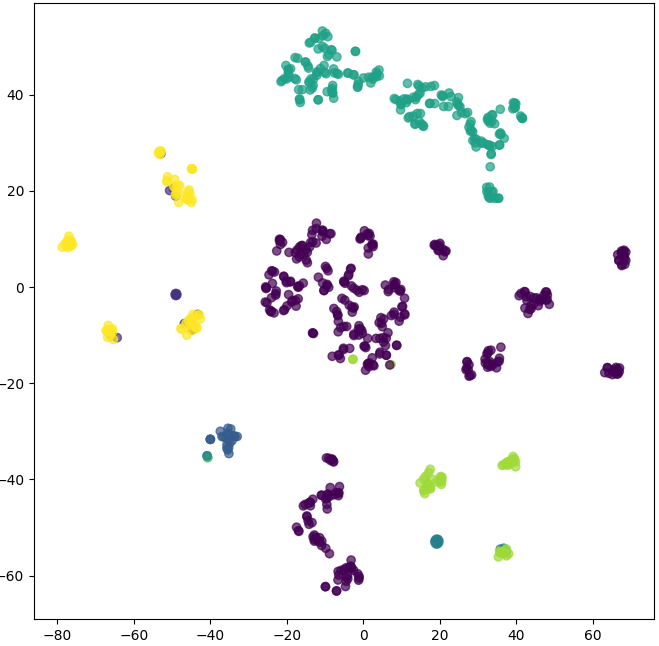}
    \caption{Hard clustering for 2 rules}
    \label{fig:2rules-a}
  \end{subfigure}
    \begin{subfigure}{0.33\textwidth}
    \centering
    \includegraphics[width=\linewidth]{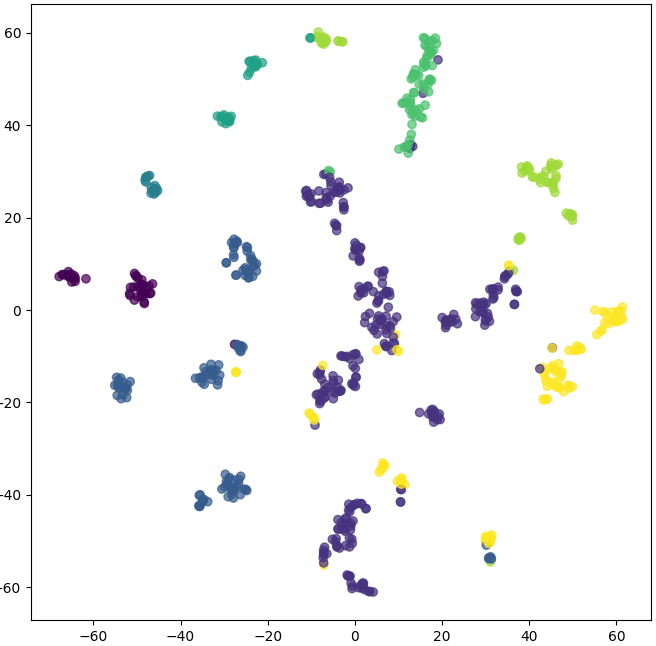}
    \caption{Hard clustering performance for 4 rules}
    \label{fig:4rules-a}
  \end{subfigure}
  
  \begin{subfigure}{0.33\textwidth}
    \centering
    \includegraphics[width=\textwidth]{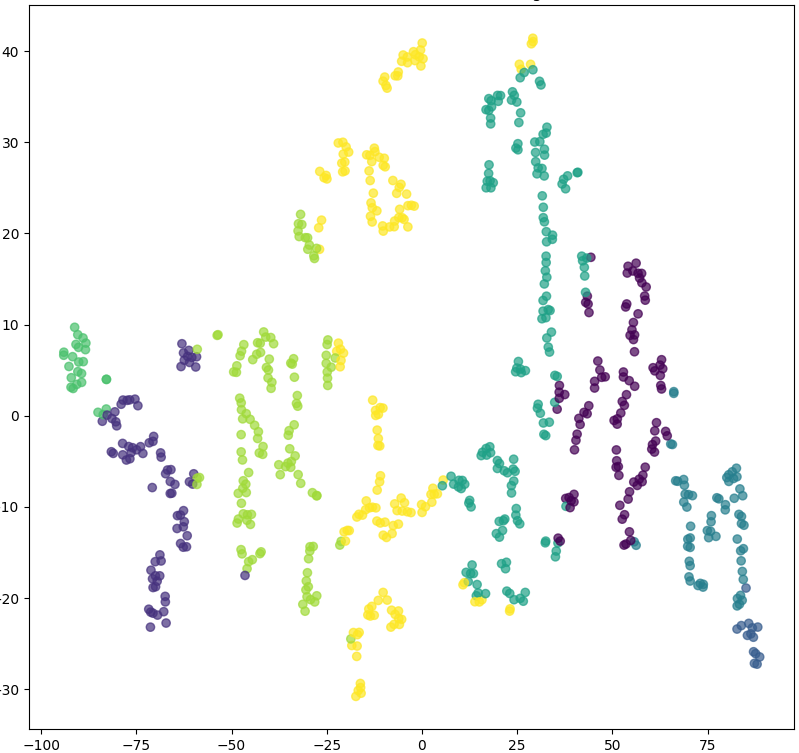}
    \caption{ Soft clustering  without rules}
    \label{fig:short-b}
  \end{subfigure} 
  \begin{subfigure}{0.33\textwidth}
    \centering
    \includegraphics[width=\textwidth]{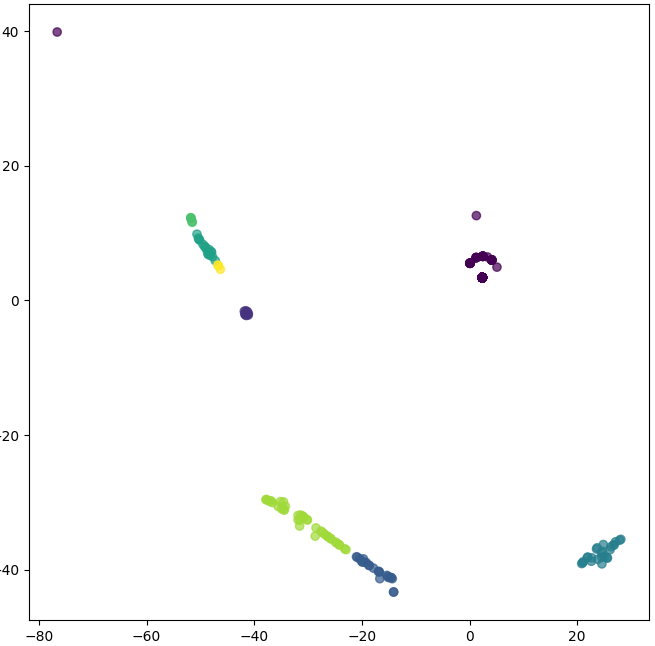}
    \caption{Soft clustering for 2 rules}
    \label{fig:2rules-b}
  \end{subfigure}
  \begin{subfigure}{0.33\textwidth}
    \centering
    \includegraphics[width=\linewidth]{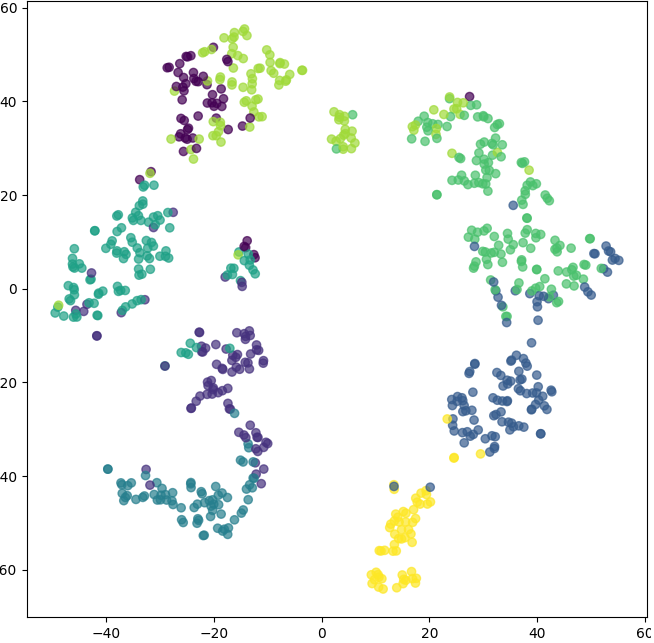}
    \caption{Soft clustering for 4 rules}
    \label{fig:4rules-b}
  \end{subfigure}
  
  \caption{Aircraft clustering performance demonstrated through t-SNE visualizations with varying physical rules and clustering methods.}
  \label{fig:aircraft-comparison}
\end{figure*}



\begin{figure*}[!ht]
   \centering
  \begin{subfigure}{0.44\textwidth}
    \centering
    \includegraphics[width=\linewidth]{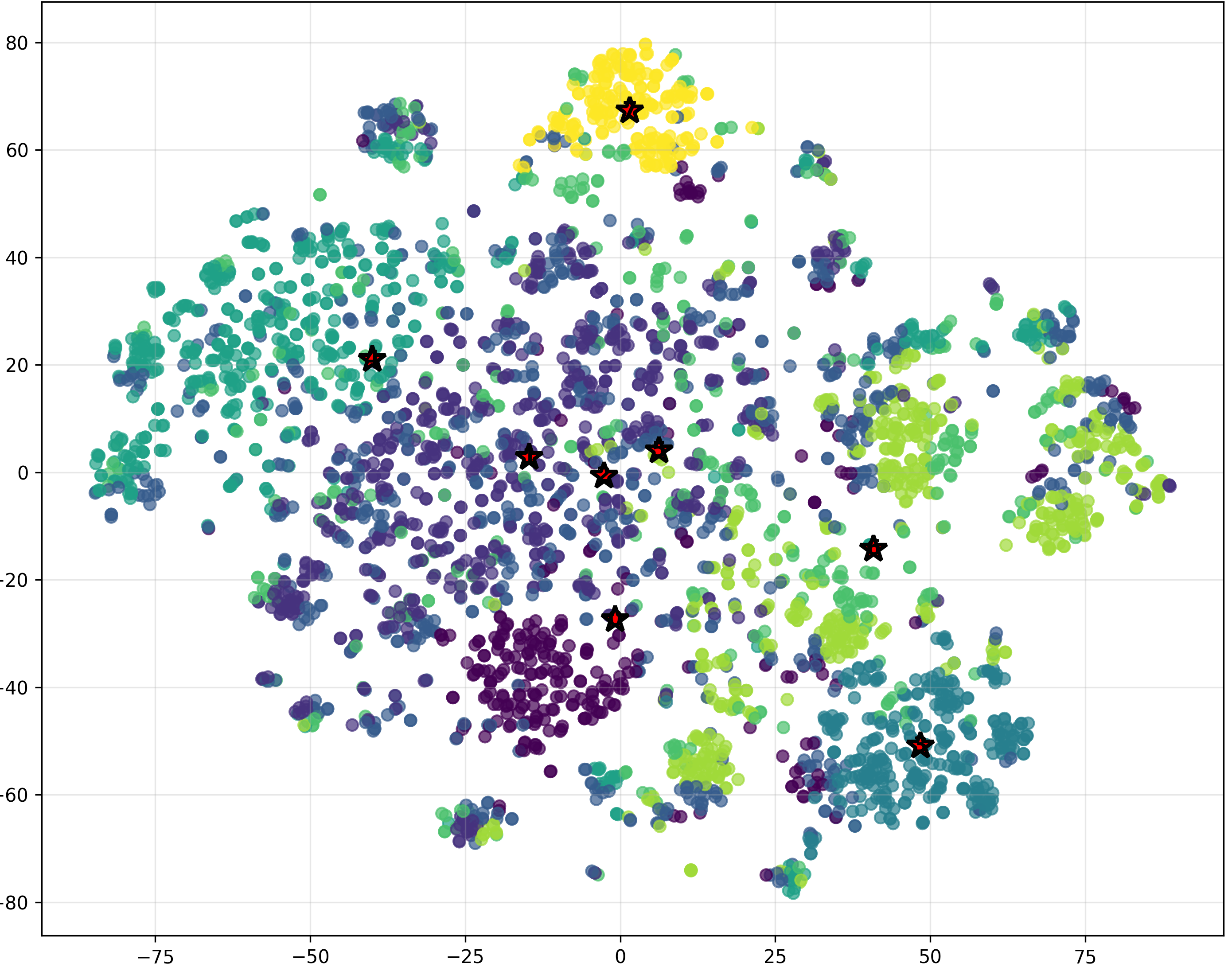}
    \caption{Without physical rules}
    \label{fig:car-hard-without-rules}
  \end{subfigure}
  \begin{subfigure}{0.44\textwidth}
    \centering
    \includegraphics[width=\linewidth]{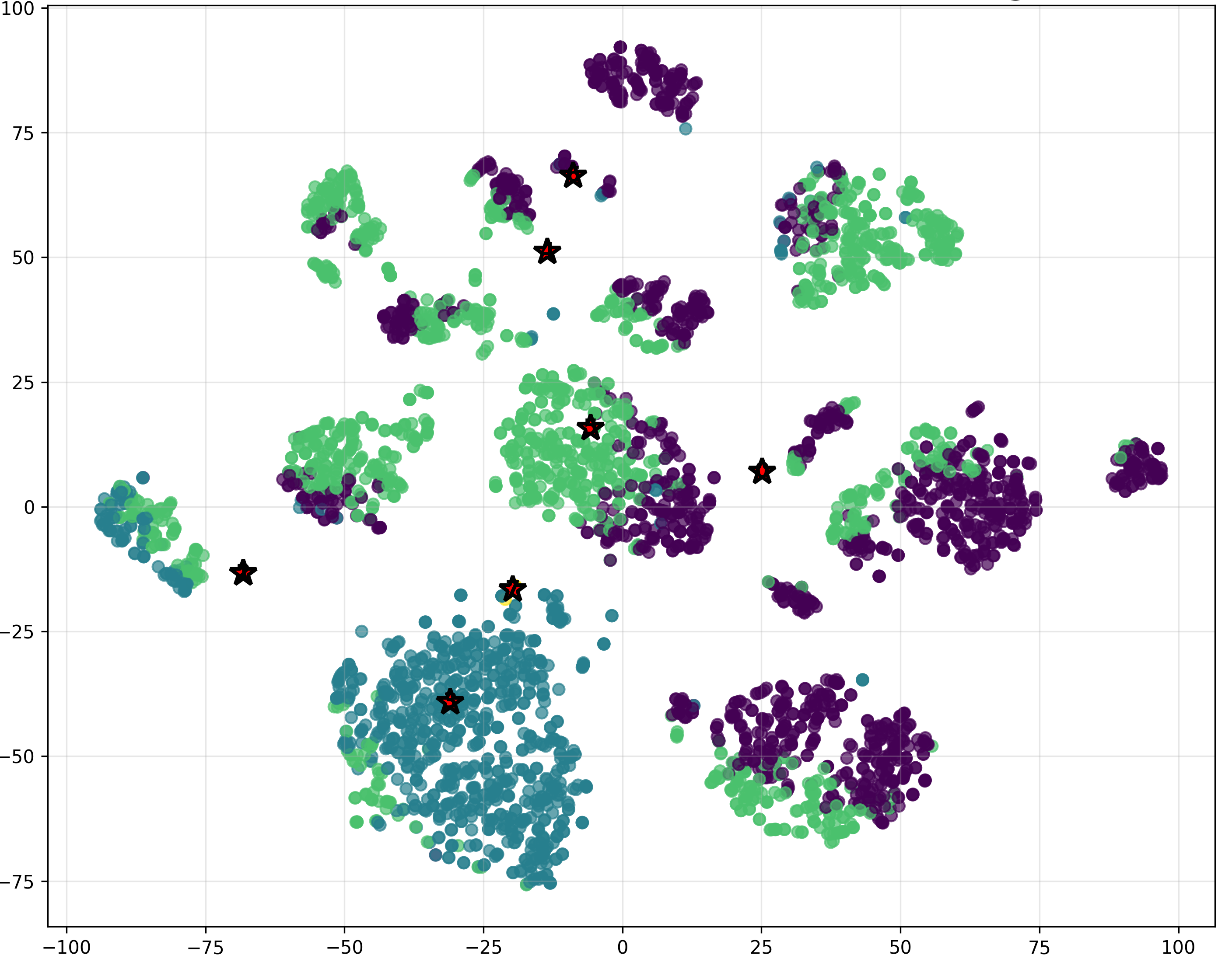}
    \caption{With 4 physical rules}
    \label{fig:car-hard-with-rules}
  \end{subfigure}
  \caption{Car Hard Clustering t-sne visualization}  
  \label{fig:car-hard-clustering}
\end{figure*}


%% file: sec/5_results_analysis.tex
\section{Results and Analysis}
\label{sec:results}

We assess DART-VAE in the aircraft and automotive domains through extensive experimental setups to illustrate the incremental advantages of including domain constraints.  All experiments employ appropriate statistical validation and adhere to the implementation specifics outlined in Section~\ref {sec:dataset}.

\subsection{Aircraft Domain Clustering Results}

\subsubsection{Quantitative Performance Analysis}

Table~\ref{tab:aircraft_clustering_combined} breaks down comprehensive aircraft clustering outcomes across various rule configurations and techniques.  The methodical advancement from baseline to rule-based grouping reveals distinct performance variation contingent upon the complexity of constraints. \textbf{Hard Clustering Performance}: The 2-rule hard configuration achieves remarkable quantitative measures, evidenced by a Silhouette Score of 0.7109 and a robust Calinski-Harabasz Score of 16,325.64, reflecting a 405\% enhancement over the baseline hard clustering score of 0.1406.  However, the 4-rule hard configuration exhibits diminished traditional metrics (Silhouette: 0.3325) while achieving improved Davies-Bouldin performance (0.9147), which means greater cluster compactness despite lowered separation scores. \textbf{Soft Clustering Performance}: Fuzzy C-means with 2-rule configuration achieves remarkable performance with Fuzzy Partition Coefficient of 0.9765 and minimal Fuzzy Partition Entropy of 0.0474, demonstrating optimal cluster separation. Table~\ref{tab:aircraft_clustering_combined} shows the fuzzy-specific metrics where the 2-rule configuration excels with high certainty in cluster assignments and Average Membership Strength of 0.9850. The 4-rule soft configuration achieves Fuzzy Partition Coefficient of 0.4736 and Fuzzy Partition Entropy of 1.1456, with Average Membership Strength of 0.6184, trading traditional fuzzy metrics for comprehensive constraint coverage.

\begin{table*}[!htbp]
\centering
\caption{Aircraft clustering performance comparison (K-means vs. Fuzzy C-means) across rule configurations}
\label{tab:aircraft_clustering_combined}
\begin{tabular}{l|cccc|cccc}
\hline
\multirow{2}{*}{Configuration} 
& \multicolumn{4}{c|}{Hard Clustering (K-means)} 
& \multicolumn{4}{c}{Rule Violations (Hard)} \\
& SS & DB & CH & &Stealth & UAV & Mission & Semantic \\
\hline
Baseline   & 0.1406 & 1.5167 & 174.37   & -- & --  & --  & --  \\
2-Rule     & 0.7109 & 1.0666 & 16325.64 & & 85 & 0   & --  & --  \\
4-Rule     & 0.3325 & 0.9147 & 794.43   & & 85 & 0   & 401 & 103 \\
\hline
\multirow{2}{*}{Configuration} 
& \multicolumn{4}{c|}{Soft Clustering (Fuzzy C-means)} 
& \multicolumn{4}{c}{Rule Violations (Soft)} \\
& FPC & FPE & MS & FS & Stealth & UAV & Mission & Semantic \\
\hline
Baseline   & 0.5401 & 0.9775 & 0.396  & --     & -- & --  & --  & --  \\
2-Rule     & 0.9765 & 0.0474 & 0.9850 & --     & 85 & 214 & --  & --  \\
4-Rule     & 0.4736 & 1.1456 & 0.6184 & 0.1960 & 85 & 90  & 533 & 0   \\
\hline
\multicolumn{9}{l}{\footnotesize SS: Silhouette Score, DB: Davies-Bouldin Score, CH: Calinski-Harabasz Score,} \\
\multicolumn{9}{l}{\footnotesize FPC: Fuzzy Partition Coefficient, FPE: Fuzzy Partition Entropy, 
MS: Average Membership Strength, FS: Fuzzy Silhouette Index}
\end{tabular}
\end{table*}

\begin{table*}[!ht]
\centering
\caption{Automotive clustering performance comparison (K-means vs. Fuzzy C-means)}
\label{tab:automotive_clustering_combined}
\begin{tabular}{lcccc}
\hline
 & FPC & FPE & MS & FS \\
\hline
Baseline (Soft)    & 0.125 & 2.07 & 0.125  & 0.167 \\
Rule-Guided (Soft) & 0.125 & 2.08 & 0.125  & 0.189 \\
 & SS & DB & CH & \\
Baseline (Hard)    & 0.054  & 3.27  & 173.8   & --     \\
Rule-Guided (Hard) & 0.139  & 0.92  & 12234.9 & --     \\
\hline
\multicolumn{5}{l}{\textbf{Rule Violations (Soft vs. Hard, Rule-Guided)}} \\
Metric      & Soft & Hard &   &   \\
Body        & 0.19 & 3.01 &   &   \\
Performance & 3.60 & 4.83 &   &   \\
Size        & 2.00 & 3.92 &   &   \\
Luxury      & 2.72 & 4.53 &   &   \\
\hline
\multicolumn{5}{l}{\footnotesize Total violations: Soft  Rule-Guided = 8.51; Hard Rule-Guided = 16.29} \\
\end{tabular}
\end{table*}

\subsubsection{Rule Violation Analysis}

\textbf{Stealth Consistency Constraints}: Violations of stealth technology persist at a consistent rate of 85 across all configurations, signifying structural issues within the stealth aircraft category that surpass clustering algorithms. This ongoing violation pattern indicates intrinsic data complexity, wherein stealth qualities do not fully correspond with other operational features. \textbf{UAV Operational Separation}: Hard clustering with rule refinement provides perfect UAV separation (0 violations) by post-processing optimization, while soft clustering keeps 90-214 violations but accepts boundary scenarios when UAV characteristics overlap with manned aircraft traits. With increasing rule complexity, constraint management improves from 214 UAV violations in the 2-rule soft setup to 90 in the 4-rule version. \textbf{Mission and Semantic Coherence}: The four-rule configurations impose constraints related to mission type and semantic coherence, resulting in 401 to 533 mission violations and 0 to 103 semantic violations. While these increase overall constraint violations, they provide fine-grained operational classification essential for military applications.

\subsubsection{Visual Clustering Analysis}

Figure~\ref{fig:aircraft-comparison} illustrates t-SNE visualizations that depict the evolution of clustering across various rule configurations.  The 2-rule guided clustering (Figure~\ref{fig:2rules-a} and~\ref{fig:2rules-b}) demonstrates effective cluster separation, resulting in well-defined operational groupings.  The configurations with four rules (Figure~\ref{fig:4rules-a} and~\ref{fig:4rules-b}) demonstrate more intricate boundaries that indicate a higher level of constraint complexity, whereas Figure~\ref{fig:2rules_aircraft} presents the baseline performance in the absence of rule guidance.

\textbf{Cluster Coherence}: Rule-guided clustering efficiently creates operational coherence, as stealth platforms (F-22, Su-57, F-117) cluster according to technological uniformity, UAVs maintain clear operational distinctions from manned aircraft, and transport aircraft (C-130, C-2) are differentiated from combat platforms (A-10).  This indicates a significant shift from visual similarity to functional effectiveness.

\subsection{Automotive Domain Clustering Results}

\subsubsection{Quantitative Performance Analysis}

Table~\ref{tab:automotive_clustering_combined} illustrates the performance of hard clustering in the automotive sector at various levels of rule integration. Rule-based hard clustering demonstrates significant enhancements: The Silhouette Score rises from 0.0543 (baseline) to 0.1393 (156\% enhancement), while the Calinski-Harabasz Score escalates from 173.82 to 12,234.91 (6,941\% enhancement), signifying significantly enhanced cluster separation and compactness.

\subsubsection{Rule Violation Analysis}

\textbf{Body Style Coherence}: The use of rule-guided clustering results in differing degrees of Body Style/Segment violations according on the clustering algorithm utilized. Tables~\ref{tab:automotive_clustering_combined} illustrate that fuzzy C-means clustering exhibits superior constraint management with merely 0.19 violations, whereas K-means clustering results in 3.01 violations. This notable disparity demonstrates fuzzy clustering's efficacy in managing the diverse classifications of body shapes in the automotive sector, including SUV, sedan, and hatchback, where automobiles may display ambiguous traits.

\textbf{Performance and Engineering Constraints}: Performance/Drivetrain Consistency is the hardest criteria for both clustering methods, with soft clustering (fuzzy C-means) obtaining 3.60 violations and hard clustering (K-means) 4.83. High violation rates reflect the car industry's sophisticated performance specs that don't match visual similarities—high-performance variations of ordinary models sometimes look identical despite having very different powertrains. Size Proportion limitations yield mild violations, with fuzzy C-means achieving 2.00 and K-means 3.92. Luxury/Performance Feature consistency provides sufficient constraint adherence, with fuzzy C-means committing 2.72 violations and K-means 4.53.

\textbf{Algorithm Comparison}: The experimental results indicate complementary benefits among clustering methodologies. K-means demonstrates superior mathematical clustering effectiveness, as indicated by a higher Silhouette Score (0.1393 compared to 0.0100) and better cluster separation metrics, making it suitable for applications that prioritize geometric cluster quality. Fuzzy C-means shows improved adherence to domain constraints, with 48\% fewer rule violations (8.51 versus 16.29), making it more suitable for applications that require semantic coherence over mathematical optimization. Figure~\ref{fig:car-hard-clustering} illustrates that rule-guided K-means clustering (Figure~\ref{fig:car-hard-with-rules}) achieves remarkable separation with 8 distinct clusters, whereas baseline clustering (Figure~\ref{fig:car-hard-without-rules}) exhibits considerable mixing.  

\subsection{Qualitative Clustering Improvements} Table~\ref{tab:improvements} demonstrates specific clustering improvements through rule integration. Critical operational separations achieved include:  \textbf{Military Applications}: MQ-9 Reaper UAVs isolated from F-18 fighters, stealth aircraft (F-22, Su-57) unified by technological consistency, and transport aircraft (C-130, C-2) separated from combat platforms.  \textbf{Automotive Applications}: BMW M3 performance vehicles separated from economy cars, Ferrari luxury sports cars grouped appropriately, and Range Rover SUVs clustered by body style coherence. These improvements represent fundamental advances from appearance-based to function-based clustering, critical for domain expert applications where operational semantics transcend visual similarity.

\section{Threats to Validity} While the proposed framework demonstrates the potential of incorporating explicit domain rules into clustering, several threats to validity remain. First, the rules themselves are generated by large language models (LLMs) and subsequently formatted as structured knowledge graphs. This process introduces the risk of hallucinations, where the LLM may produce inaccurate or spurious rules. Such negative rules can inadvertently bias the clustering process, leading to distortions rather than improvements in cluster quality. Second, the reliance on LLM-generated rules makes the framework sensitive to overfitting. When too many rules are imposed simultaneously, the latent space may become overly constrained, forcing the model to adhere to rigid relationships at the expense of generalizability. This effect was particularly evident when scaling from two-rule to four-rule configurations, where clustering metrics showed diminishing returns despite improved constraint enforcement. Third, scaling to complex domains presents a fundamental challenge. As the number of rules, modalities, and semantic categories increases, the computational overhead and difficulty of maintaining consistent enforcement escalate. In such cases, rule conflicts and inconsistencies may proliferate, complicating both training and interpretability. Without careful curation and validation of rule sets, domain expansion risks undermining the stability and reliability of the method. Finally, the evaluation relies on datasets where rule definitions are relatively well aligned with domain expertise (e.g., aircraft doctrine, automotive market segmentation). For less formalized or more ambiguous domains, the suitability of LLM-derived rules remains uncertain, limiting the external validity of the approach.

%% file: sec/6_conclusion.tex
\section{Conclusion}
\label{sec:conclusion}
We proposed DART-VAE, a multimodal clustering framework that elevates LLM-generated rules and knowledge-graph constraints to first-class learning signals. By embedding rules directly into representation learning, the method balances visual coherence and operational semantics, producing clusters that are both interpretable and quantitatively robust. Results in aircraft and automotive domains show that rule-guided clustering achieves clearer functional separation than purely visual baselines. However, reliance on LLM-generated rules introduces risks of hallucination and inconsistency, and applying too many constraints can lead to overfitting. Scaling to complex domains with large rule sets remains challenging. Despite these limitations, DART-VAE demonstrates the potential of rule-informed clustering as a principled step toward interpretable and domain-aligned AI.